\documentclass[twoside,11pt]{article}

\usepackage{graphicx}
\usepackage{jmlr2e}

\firstpageno{1}

\begin{document}

\title{openXBOW -- Introducing the Passau Open-Source Crossmodal Bag-of-Words Toolkit}

\author{\name Maximilian Schmitt \email maximilian.schmitt@uni-passau.de \\
        \name Bj\"orn W.\ Schuller\,\thanks{B.\,W.\ Schuller is also with the Department of Computing, Imperial College London, UK.} \email bjoern.schuller@uni-passau.de \\
        \addr Chair of Complex and Intelligent Systems\\
        University of Passau\\
        Passau, Germany}

% LATER UNCOMMENT ALSO IN STYLE FILE
%\editor{Leslie Pack Kaelbling}

\maketitle

\begin{abstract}%   <- trailing '%' for backward compatibility of .sty file
We introduce \textsc{openXBOW}, an open-source toolkit for the generation of bag-of-words (BoW) representations from multimodal input. In the BoW principle, word histograms were first used as features in document classification, but the idea was and can easily be adapted to, e.\,g., acoustic or visual low-level descriptors, introducing a prior step of vector quantisation.
The \textsc{openXBOW} toolkit supports arbitrary numeric input features and text input and concatenates computed subbags to a final bag. It provides a variety of extensions and options. To our knowledge, \textsc{openXBOW} is the first publicly available toolkit for the generation of crossmodal bags-of-words. The capabilities of the tool are exemplified in two sample scenarios: time-continuous speech-based emotion recognition and sentiment analysis in tweets where improved results over other feature representation forms were observed.
\end{abstract}

\begin{keywords}
  bag-of-words, multimodal signal processing, histogram feature representations
\end{keywords}

\section{Introduction}

The bag-of-words (BoW) principle is a common practice in natural language processing (NLP) \citep{Woellmer12-FAA,Weninger13-WTF}. In this method, \textit{word histograms} are generated, i.\,e., within a text document, the frequencies of words from a dictionary are counted. The resulting word-frequency vector is then input to a classifier, such as \textit{na\"ive Bayes} or a  \textit{support vector machine} (SVM), i.\,e., machine learning schemes which are known to cope well with possibly irrelevant features and large, yet sparse feature vectors.

One major drawback of the BoW approach is that the order of the words in a document, which often implies important information, is not taken into account. In order to overcome this problem, \textit{n-grams} have been employed, where sequences of $n$ words or characters (\textit{n-character-grams}) are counted instead of single words \citep{wallach2006topic,Schuller09-TGV}.

BoW has been adopted by the visual community, where it is known under the name bag-of-visual-words (BoVW) \citep{fei2005bayesian,sivic2005discovering}. Instead of lexical words, local image features are extracted from an image and then their general distribution is modelled by a histogram according to a learnt codebook, which substitutes the dictionary from BoW.

In recent years, the principle has also been employed successfully in the field of audio classification, where it is known under the term bag-of-audio-words (BoAW). Acoustic low-level descriptors (LLDs), such as mel-frequency cepstral coefficients (MFCC), are extracted from the audio signal, then, the LLD vectors from single frames are quantised according to a codebook \citep{pancoast2012bag}. This codebook can be the result of either k-means clustering \citep{Pokorny15-DON} or random sampling of LLD vectors \citep{RawatSBDWM13}. Other approaches employ expectation maximisation (EM) clustering, which leads to a soft vector quantisation step \citep{Grzeszick2015-TAW}. A histogram finally describes the distribution of the codebook vectors over the whole audio signal or one segment.
Major applications of BoAW are acoustic event detection and multimedia event detection \citep{liu2010coherent,pancoast2012bag,RawatSBDWM13,plinge2014bag,Lim15-RSE} but they have also been successfully used in music information retrieval \citep{RileyHG08} and emotion recognition from speech \citep{Pokorny15-DON}.

In this contribution, we introduce the first open-source toolkit for the generation of BoW representations across modalities, thus named `\textsc{openXBOW}' (``open cross-BoW''). The motivation behind \textsc{openXBOW} is to ease the generation of a fused BoW-based representation from different modalities. These modalities can be the audio or the visual domain, providing numeric LLDs, and written documents or transcriptions of speech, providing text. However, in principle arbitrary other types of modalities ``$x$'' can be thought off, such as stemming from physiological measurement or feature streams as used in brain computing, etc. As a placeholder for arbitrary modalities, we thus introduce the notion of bag-of-x-words or BoXW for short. In case of multimodal or `crossmodal' usage, the output of the toolkit is a concatenated feature vector consisting of histogram representations per modality or combinations of these. Multimodal BoW have already been employed, e.\,g., for depression monitoring in \citep{joshi2013multimodal}, exploiting both the audio and the video domain. 

\textsc{openXBOW} provides a multitude of options, e.\,g., different modes of vector quantisation, codebook generation, term frequency weighting, and methods known from natural language processing to process the textual features. To the knowledge of the authors, such a toolkit has not been published, so far, whereas there are already some libraries implementing BoVW, such as \textit{DBoW2} \citep{galvez2012bags}.

In the next section, we give an overview of the \textsc{openXBOW} tool, its structure and its options. In section~\ref{sec:experiments}, we give results from two exemplary applications of the tool. We conclude and give an outlook on future developments in \textsc{openXBOW} in the final section~\ref{sec:conclusionsandoutlook}.

\section{Overview}
\label{sec:overview}

\textsc{openXBOW} is implemented in Java and can thus be used on any platform. It has been published on GitHub as a public repository\footnote{https://github.com/openXBOW/openXBOW}, including both the source code and a compiled jar file for those who do not have a Java Development Kit installed. The software and the source code can be freely used by the research community for non-commercial purposes.

\textsc{openXBOW} supports three different file formats for input of LLDs and text and the output of the BoW feature vector:
\begin{itemize}
    \item ARFF (Attribute-Relation File Format), used in the machine learning software \textit{Weka} \citep{hall2009weka}
    \item CSV (Comma separated values), with separator semicolon (;)
    \item LIBSVM file format, used in LIBSVM \citep{CC01a} and LIBLINEAR \citep{fan2008liblinear} (only output)
\end{itemize}

\begin{figure}
    \begin{minipage}{\textwidth}
        \includegraphics[width=\textwidth]{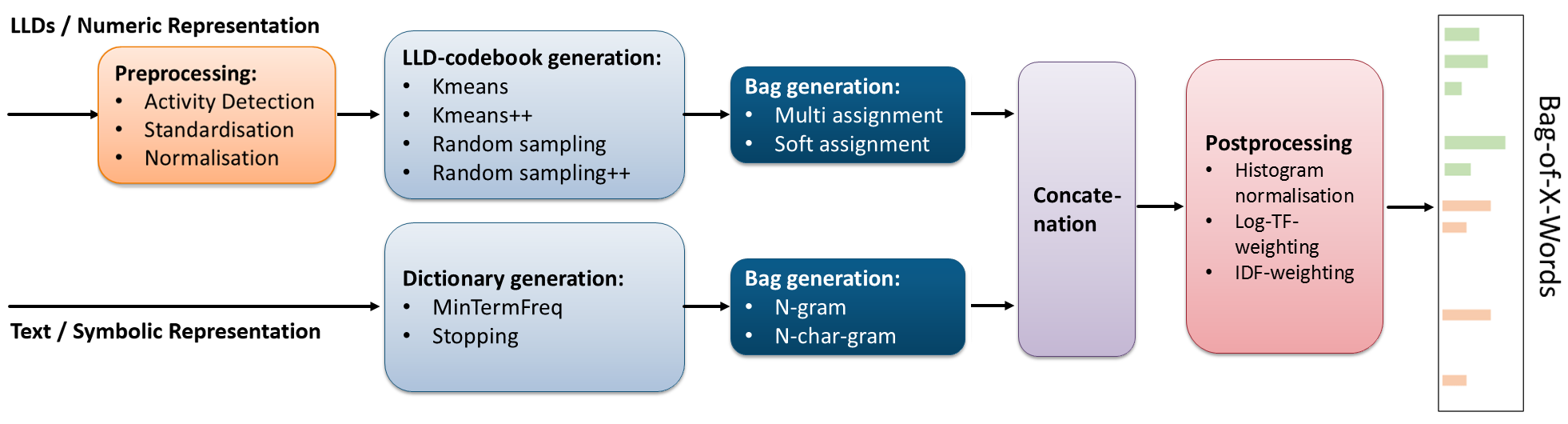}
        \caption{Overview of the workflow of \textsc{openXBOW}.}
        \label{fig:overview}
    \end{minipage}
\end{figure}

The input is processed by \textsc{openXBOW} in the way shown in figure~\ref{fig:overview}. First, there is an optional preprocessing stage, where input LLDs with a low activity can be excluded from further processing. This is especially relevant in the field of speech recognition, where LLDs at instants of time without voice activity should not be considered as they describe only the background noise.

In case feature types with (significantly) different ranges of values are combined in one BoW, normalisation or standardisation of the LLDs is essential. Both options are available, the corresponding parameters are stored in the codebook file, so that they can be applied in the same way to a given test file (online approach).

All options with a corresponding help text are displayed in the command line window if \textsc{openXBOW} is started without any arguments or the argument (-h), e.\,g.,
\begin{verbatim}
java -jar openXBOW.jar -h
\end{verbatim}
All configurations are made in the command line call. 

For codebook generation, there are four different methods available at the time; \textit{random~sampling} means that the codebook vectors are picked randomly from the input LLDs, whereas \textit{random~sampling++} favours far-off vectors as proposed in \citep{arthur2007k} as an improved initialisation step for \textit{kmeans}, called in that case \textit{kmeans++}. For the latter two methods, up to 500 updates of cluster centroids and cluster assignments are executed after initialisation.

In case of nominal class labels, the codebook generation can also be done in a supervised manner, learning a codebook from all LLDs in one class separately, first, and then concatenating these codebooks to form a \textit{super-codebook} \citep{Grzeszick2015-TAW}.

In case of large LLD vectors, \textit{split vector quantisation} (SVQ) can be suitable, as proposed in \citep{Pokorny15-DON}, where subvectors are first quantised and then the indexes of the quantised subvectors are processed in the usual scheme.
However, it is also possible to split the input LLDs manually into well-defined subvectors in order to generate different codebooks for different feature types.

A way of doing \textit{soft} vector quantisation without EM-clustering, is applying \textit{Gaussian encoding} to the term frequencies (TF), i.\,e., the number of occurrences of each word from the codebook \citep{pancoast2014softening}. The TF is then weighted in each word assignment by the distance to the word in the codebook. This can be especially useful in combination with \textit{multiple assignments}, where not only the closest word from the codebook is considered but a certain number $N_a$ of closest words.

For the text domain, standard processing techniques are available, such as \textit{stopping}, \textit{n-grams}, and \textit{n-character-grams}.

Finally, the BoW representations of different domains are fused into one feature vector. For post-processing, logarithmic TF-weighting ($\mathrm{TF_{log}}=\mathrm{lg}(\mathrm{TF}+1)$), and inverse document-frequency (IDF) weighting ($\mathrm{TF_{IDF}}=\mathrm{TF} \cdot \mathrm{lg}\frac{\mathrm{N}}{\mathrm{DF}}$, N: Number of instances, DF: Number of instances where the word is present) can be applied \citep{RileyHG08}. 

The resulting histograms can then be normalised, a step which is essential if the input instances to be classified have different sizes or if (voice or any other modality's) activity detection is used and so different segments of the input signal have different numbers of assigned words.

\section{Experiments}
\label{sec:experiments}

In this section, the usage of \textsc{openXBOW} is exemplified in two scenarios: time-continuous processing of speech input for emotion recognition, and BoW processing of tweets.

In fact, there are comparably few suitable multimodal databases with acoustic, visual, and text input available. In this introductory paper, we exemplify the principle on separate tasks for audio and text words to showcase the principle. However, results of \textsc{openXBOW} with multimodal input will be shown in our future efforts.

\subsection{Time- and value-continuous emotion recognition from speech}

Emotion recognition from speech has been conducted on the RECOLA (Remote Collaborative and Affective Interactions) corpus \citep{ringeval2013introducing}. In this database, 46 French participants have been recorded in dyadic remote collaboration for 5 minutes, each. During their collaboration, their face, speech and some physiological measures have been recorded. In our present example, we focus on the audio domain.

A gold standard in terms of \textit{arousal} and \textit{valence}, generated from the annotations by six different persons, was used as a target in our experiments.
The corpus was split into three partitions, a training partition (16 subjects) in order to learn the codebook and train the classifier, a validation partition to optimise the parameters of \textsc{openXBOW} and support vector regression (SVR) with a linear kernel and a test partition to prove the universality of the learnt regressor. The input LLDs (MFCCs 1--12 and log-energy) were extracted with our toolkit openSMILE \citep{Eyben13-RDI}.

To learn a codebook and generate a BoAW representation (\texttt{BoAW\_arousal\_train.arff}) from the training file \texttt{LLD\_train.csv}, the following options have been applied:
\begin{verbatim}
-i LLD_train.csv -o BoAW_arousal_train.arff -l arousal_train.csv -t 8.0 0.8 
-standardizeInput -size 1000 -c random++ -B codebook.txt -a 20 -log
\end{verbatim}
The option \texttt{-t 8.0 0.8} means that the sequence of input LLDs is segmented into windows of 8.0 seconds width and the hop size between two successive windows is 0.8 seconds. Labels for arousal in the file \texttt{arousal\_train.csv} must be available for exactly those instants in time (see option \texttt{-h} for information on the labels file format).

After standardisation, a codebook of 1\,000 words is generated by \textit{random~sampling++} and stored in the file \texttt{codebook.txt} together with information on the parameters of standardisation (mean, and standard deviation per LLD). Each LLD is then assigned to the 20 closest (\texttt{-a 20}) words in the codebook. Finally, the number TFs are compressed applying logarithmic TF-weighting. Also, this information is stored in the codebook files, as it must be used in the same way with the validation and test instances. The order of the options does not have any effect.

After the optimum configuration of BoAW has been found (let us assume that it is the one described above), the BoAW can be generated from the validation (and also test) files (\texttt{\-valid}) in the following way:
\begin{verbatim}
-i LLD_valid.csv -o BoAW_arousal_valid.arff -l arousal_valid.csv -t 8.0 0.04 
-b codebook.txt -a 20
\end{verbatim}
The codebook is loaded with the command \texttt{-b}. Please note that, the file \texttt{codebook.txt} implies also the standardisation and the log-TF-weighting.
The hop size is chosen differently for the validation partition; it is only 0.04 seconds (\texttt{-t 8.0 0.04}) which is not an issue if the corresponding labels are included in the labels file (\texttt{arousal\_valid.csv}).

Table~\ref{tab:resultsRECOLA} shows the results of BoAW for emotion recognition compared to the baseline of the AVEC 2016 challenge at ACM Multimedia 2016 \citep{valstar2016avec}, which is also carried out on RECOLA under the same conditions. However, the number of recordings in the respective training, validation and test set in the AVEC 2016 challenge were smaller than the ones we used.

The optimum window size for the prediction of valence is 10.0 seconds, compared to 8.0 seconds for arousal. All labels have been shifted to the front in time in order to compensate the natural delay between the shown emotion of the subject and the reaction of the annotator. A shift of 4.0 seconds prove to be an optimum. The complexity of SVR has been optimised for each configuration. As a metric for evaluation, the concordance correlation coefficient (CCC) is used, which takes also scaling of the outputs into account, compared to the linear correlation coefficient. Note that, the CCC is also used as competition measure in the AVEC 2016 competition.

\begin{table}[h]
    \caption{\label{tab:resultsRECOLA} {\it Performance of speech-based emotion recognition on RECOLA using BoAW compared to the baseline of AVEC 2016. Shown are results on the official Valid(ation) and Test sets. Results printed in \textbf{bold} are statistically significantly better than the baseline (level of significance: 0.01)}}
    \centerline{
        \footnotesize
        \begin{tabular}{|l|r|r|r|r|}
            \hline
            Model & \multicolumn{4}{c|}{CCC} \\
                  & \multicolumn{2}{c|}{Arousal} & \multicolumn{2}{c|}{Valence} \\
                  & Valid & Test & Valid & Test \\
            \hline \hline
            Baseline AVEC 2016 (audio only) & .796 & .648  & .455 & .375 \\
            \hline
            \textsc{openXBOW} (BoAW)        & .793 & \textbf{.753} & \textbf{.550} & \textbf{.430} \\
            \hline
        \end{tabular}
    }
\end{table}

It can be clearly seen that, our approach significantly outperforms the baseline on the validation and test sets, except for the performance for arousal on the validation partition, where a similar result is achieved.

\subsection{Twitter sentiment analysis}

Our second experiment shows obtainable performances on written text on another well-defined task.
A large corpus of short messages from the social network \textit{Twitter}\footnote{http://twitter.com}, known as `tweets', in English language is provided by \textit{Thinknook}\footnote{http://thinknook.com/twitter-sentiment-analysis-training-corpus-dataset-2012-09-22/}. This dataset has been collected by the \textit{University of Michigan} and \textit{Niek Sanders}. It includes 1\,578\,627 tweets with a 2-class annotation of either positive or negative sentiment. An accuracy of 75\,\% is reported as state-of-the-art.

In order to train a dictionary and create a BoW representation from the training set \texttt{senti-train.csv}, the following command line arguments can be used:

\begin{verbatim}
-i senti-train.csv -attributes ncr0 -o BoW.arff -B dictionary.txt 
-minTermFreq 1000 -maxTermFreq 30000 -nGram 2 -log -idf
\end{verbatim}
The option \texttt{-attributes} is needed if the data structure of the input is not the standard defined in the \textsc{openXBOW} help text. It specifies each input feature in the input file, i.\,e., each column in an input CSV file. \texttt{n} means that, the corresponding column (the first column here) specifies the \textit{name} (or an index) as a unique ID this line belongs to. All lines in the input with the same name are put into one bag later on. The second column in the Twitter dataset (\texttt{c}) specifies the class label. The third column (\texttt{r}, remove) can be discarded as it does not contain any relevant information, the last column of the input contains the text to be classified. The digit \texttt{0} always specifies text input, the digits \texttt{1} to \texttt{9} specify numeric input. A separate codebook is generated for every digit if at least one feature with a digit is present. More information about the input format can be found in the help text.

\texttt{-minTermFreq} and \texttt{-maxTermFreq} implement stopping as known in NLP. Thereby, either very rare words or very common words (such as `and', or `or') can be excluded from the dictionary as/if they are not likely to carry meaningful information. 2-grams (\texttt{-nGram 2}) and log-TF-IDF weighting (\texttt{-log -idf}) are also used in this sample call.

To apply the same model to the test set, the following line is used:
\begin{verbatim}
-i senti-test.csv -attributes ncr0 -o BoW.arff -b dictionary.txt -nGram 2
\end{verbatim}
Note that, the TF-weighting is stored in the dictionary and is thus not needed; the option \texttt{-nGram 2} must, however, be repeated.

In our experiments, the best results have been achieved without using (higher order) n-grams, i.\,e., simply employing uni-grams or each word by itself, respectively. We split the whole corpus into two partitions, the first 1\,000\,000 instances form the training set, the remaining 578\,627 instances form the test set. Support vector machine with linear kernel was again found to be suitable to handle the high-dimensional BoW vector.

A \textit{minimum term frequency} of 500 and a \textit{maximum term frequency} of 100\,000, leading to a dictionary size of 1\,875 terms was found suitable. With an SVM complexity of 0.1, a weighted accuracy (WA) of 77.28\,\% and an unweighted accuracy (UA) of 77.29\,\% have been achieved. These results appear to match or outperform a little bit the state-of-the-art. Even better performance might be achieved with more sophisticated approaches, such as \textit{long short-term memory recurrent neural networks} \citep{Schuller15-SAA}.

\section{Conclusions and outlook}
\label{sec:conclusionsandoutlook}

We introduced our novel \textsc{openXBOW} toolkit -- a first of its kind -- for the generation of BoW representations from multimodal symbolic (including text), but also numeric (such as audio or video feature streams) information representations. In two (monomodal) examples we showed the potential of the toolkit and the underlying BoXW principle by outperforming the state-of-the-art on a well-defined modern speech-based emotion recognition competition task. % and reaching the state-of-the-art on a textual sentiment task. 
The full potential is, however, likely to be revealed once targeting actual crossmodal tasks. Likewise, it stands to reason that in the future, the performance can be improved using input from the visual domain, the physiological domain, and the transcribed speech on the emotion recognition task considered. 

We have also shown that our toolkit already provides state-of-the-art results in text classification.

Future work on \textsc{openXBOW} will include further soft vector quantisation techniques such as using EM clustering or non-negative matrix factorisation-based soft clustering and methods taking the order of the crossmodal words into account, such as temporal augmentation \citep{Grzeszick2015-TAW} and n-grams for numeric features \citep{pancoast2013n}. Another goal is to add a graphical user interface for configuration in order to make the work more convenient.

The public repository\footnote{https://github.com/openXBOW/openXBOW} will be regularly updated. The authors kindly request for feedback.

% Acknowledgements should go at the end, before appendices and references
%\newpage %ADDED

\acks{This work has been supported by the European Union's Horizon 2020 Programme under grant agreement No.\ 645094 (IA SEWA) and the European Community's Seventh Framework Programme through the ERC Starting Grant No.\ 338164 (iHEARu).
%The research leading to these results has received funding from the European Union's Horizon 2020 Programme under grant agreements No.\ 645094 (IA SEWA) and the European Community's Seventh Framework Programme through the ERC Starting Grant No.\ 338164 (iHEARu).
%This work was partially supported by the European Union's Horizon 2020 Programme through the Innovative Action No.\ 645094 (SEWA) and the EC's 7th Framework Programme through the ERC Starting Grant No.\ 338164 (iHEARu).
}


% TODO
%\newpage

\vskip 0.2in
\bibliography{literature,Schuller}

\end{document}